\documentclass[conference]{IEEEtran}
\IEEEoverridecommandlockouts
\usepackage{cite}
\usepackage{amsmath,amssymb,amsfonts}
\usepackage{algorithmic}
\usepackage{graphicx}
\usepackage{textcomp}
\usepackage{xcolor}
\def\BibTeX{{\rm B\kern-.05em{\sc i\kern-.025em b}\kern-.08em
    T\kern-.1667em\lower.7ex\hbox{E}\kern-.125emX}}
\begin{document}

\title{Dynamic Time Slot Allocation Algorithm for Quadcopter Swarms\\
\thanks{This work has been funded by the German Research Foundation (DFG) as part of sub-project C3 within the Collaborative Research Center (CRC) 1053 – MAKI.}
}

\author{
\IEEEauthorblockN{Sharif Azem}
\IEEEauthorblockA{\textit{Self-Organizing Systems Lab} \\
\textit{Technische Universität Darmstadt}\\
Darmstadt, Germany \\
sharif.azem@stud.tu-darmstadt.de}
\and
\IEEEauthorblockN{Anam Tahir}
\IEEEauthorblockA{\textit{Self-Organizing Systems Lab} \\
\textit{Technische Universität Darmstadt}\\
Darmstadt, Germany \\
{anam.tahir@tu-darmstadt.de}}
\and
\IEEEauthorblockN{Heinz Koeppl}
\IEEEauthorblockA{\textit{Self-Organizing Systems Lab} \\
\textit{Technische Universität Darmstadt}\\
Darmstadt, Germany \\
{heinz.koeppl@tu-darmstadt.de}}
}

\maketitle

\begin{abstract}
A swarm of quadcopters can perform cooperative tasks, such as monitoring of a large area, more efficiently than a single one. 
However, to be able to successfully work together, the quadcopters must be aware of the position of the other swarm members, especially to avoid collisions.
A quadcopter can share its own position by transmitting it via radio waves and in order to allow multiple quadcopters to communicate effectively, a decentralized channel access protocol is essential. 
We propose a new dynamic channel access protocol, called Dynamic time slot allocation (DTSA), where the quadcopters share the total channel access time in a non-periodic and decentralized manner. 
Quadcopters with higher communication demands occupy more time slots than less active ones. 
Our dynamic approach allows the agents to adapt to changing swarm situations and therefore to act efficiently, as compared to the state-of-the-art periodic channel access protocol, time division multiple access (TDMA).
Along with simulations, we also do experiments using real Crazyflie quadcopters to show the improved performance of DTSA as compared to TDMA.
\end{abstract}


\section{INTRODUCTION}
\label{section:introduction}

There are many everyday applications in which multiple quadcopters have to work together to perform a complex task. 
One example would be a surveillance of an area after crisis to analyse the damage or search and rescue operation to look for survivors \cite{erdelj2017help, radoglou2020compilation, semsch2009autonomous}.  
Such tasks can be performed either in a centralised manner, where the agents\footnote{We use the terms quadcopters and agents interchangeably.} are given a predefined trajectory that they need to follow, or in a decentralised way, in which the agents just know the end destination and need to adapt the trajectory on the go. 
In the centralised case, the agents are not adaptive to the changes in the environment (or swarm).
Hence, there is a need to look into decentralised algorithms where the agents decide their next steps on the go, based on current available information.

One prominent research area in unmanned multi-agent systems is also the decentralised collision avoidance between agents, especially when the number of quadcopters is large \cite{yasin2020unmanned, sharma2020communication}.  
In order to avoid collisions, the quadcopters (at least) need to be aware of the position of others, which can be achieved by communication in a decentralised manner.
Communication between quadcopters is often done over the same frequency using the time division multiple access (TDMA) scheme  \cite{falconer1995time}. 
The available time is divided amongst the available quadcopters and they transmit their information, in a round robin manner, only in their allocated slots.
Current implementation of TDMA for quadcopters is not adaptive and does not change the time allocations during the task completion. 
The slots allocated initially are used throughout.
This is less efficient when some quadcopters are entering at a later time, and it is also not efficient for the case when some quadcopters leave in between.


Another weakness of TDMA is allocating slots to agents which do not change their positions too often or at all. 
TDMA also results in a risky flight when the number of quadcopters is large, since the time to transmit comes very late for each quadcopter, making its last transmitted information very outdated. 
In this work, we refer to the physical collision of the quadcopters during flight, unless stated otherwise.
Apart from collisions, the completion time of a task using TDMA with real quadcopters is higher as compared to an adaptive time allocation strategy, because of the outdated information of other agents.
We will also show this with our results.

In this work, we did not focus on other channel access schemes like frequency division multiple access (FDMA) \cite{sari1997analysis} and code division multiple access (CDMA) \cite{zigangirov2004theory}, since 
they would require additional hardware resources and higher computational power. Another option would be to use the carrier sense multiple access (CSMA) scheme \cite{csma}. However, since each moving agent must transmit its position during the flight, one agent could occupy the channel for a very long time, before releasing it to the other agents. In addition,  packet collisions are possible with CSMA, which can lead to collision between quadcopters because they will be less aware of the accurate position of each other. Instead of dealing with packet collisions, one can avoid them by dividing the medium into time slots and let the agents cooperate by sharing them according to their needs while considering the flight performance of each agent.
Hence, there is a need for a more adaptive time division access algorithm that can be implemented on the current available hardware, which we propose in this work. 
Our main contributions are:
\begin{itemize}
    \item We propose a dynamic time slot allocation algorithm, DTSA, to be used for communication in quadcopters.
    \item Our proposed algorithm is decentralised and adapts to the flight behaviour of other quadcopters in the swarm.
    \item We validate our results not only through simulations, but also through experiments on Crazyflie quadcopters, for which such a scheme has not been proposed yet.
\end{itemize}

The paper is structured as follows: In section \ref{section:system model}, we describe the system model and the need for adaptiveness in channel access. In section \ref{section:approach}, we propose our time division access scheme, dynamic time slot allocation (DTSA). In section \ref{section:results}, we prove the effectiveness of our approach using simulations and experiments on real Crazyflie quadcopters \cite{crazyflie} and we conclude in section \ref{section:conclusion}.
\section{System Model}
\label{section:system model}
In this section we will explain the general settings of our system, mostly from the simulations point of view. 
In Section \ref{subsection:experiments} we give details of how these settings are realised for real-world experiments on the Crazyflie quadcopters.
We consider $N$ quadcopters, or agents, which are located in a certain area and need to perform some task together. 
Initially the quadcopters are located at certain positions in a room such that they form a circle or an ellipsoid around the center of the room. We consider the center of the room to be at the coordinates $(x=0,y=0,z=0)$.
These shapes are chosen in order to incorporate more quadcopters into the Crazyflie experiments.
The system model is such that each quadcopter has to exchange its position with the agent on its opposite side, while avoiding collisions with all other agents in the area. 
This could be useful in scenarios where multiple quadcopters are trying to do surveillance or search-rescue in a large area from different angles, without having too much overlap.

Each agent, $a$, can determine its own velocity, $\mathbf{v}(a)$, and global position, $\mathbf{p}(a)$\footnote{Throughout this work we use bold and small letters to represent vectors.}.
This information is then broadcasted to all other agents in communication range to be used for trajectory planning with collision avoidance.  
In order to avoid interference of signals from different agents, the transmission medium is divided into time slots and agents can transmit only in their non-overlapping allocated time slot, $t_s$. 
This scheme is known as TDMA. 
The current quadcopter to quadcopter communication method of the Crazyflie includes broadcasting, which allows one Crazyflie to send a message to all other Crazyflies in communication range. However, TDMA is not yet implemented \cite{p2p_comm}.
Therefore, we implemented a simple, yet sufficient TDMA algorithm for the Crazyflie quadcopters.

The implementation of the TDMA scheme works such that each agent is assigned the time slot, $t_{s}$, that is equal to its own id. These ids are allocated initially to all agents and then maintained till the end of the task.
Thus, the agent with id $1$ transmits first, then agents $2$, $3$, \ldots, \,$N$ follow. After time slot $N$ a new round begins starting from agent $1$.

In the TDMA scheme, the scheduling procedure works in a round-robin fashion and is periodic, predefined and static. 
This strict scheduling procedure, where all the agents get access to the communication channel every time means that the update time, for each agent, remains constant during the whole operation, and that it increases with the swarm size. 
We define the update time in TDMA, $\hat t_{u,a}$, as the time agent $a$ has to wait before it gets a time slot and can send a packet again. It can be written as:  
\begin{equation}
    \hat t_{u,a} = N \, t_{s}, \quad \quad a \in A
    \label{eq:update_time_tdma}
\end{equation}
where, $A = \{1, \ldots, N\}$ is the set of all agents.
One major drawback of using such a scheme is seen in the scenario when some agents are already at their final target position while other agents are still flying. 
For example, consider a swarm with $4$ agents where only $2$ agents are still flying while the other $2$ have already successfully exchanged positions and are now just hovering. 
Ideally, only the $2$ flying agents which are generating new information (in our case positions) should be transmitting,
but in TDMA the $2$ hovering agents also keep transmitting redundant information in their allocated time slots, periodically.
Hence, there is a need for a dynamic approach for the quadcopters to communicate, which we propose in the next section.

\section{Our Approach}
\label{section:approach}


The main objective of our DTSA algorithm is to allocate time slots only to the senders that generate new information. 
Therefore, the algorithm updates a set of potential senders, $A_{p} \in A$, for each iteration $i$. An iteration is the time it takes to estimate the global position of each swarm member, select the next sender and transmit a data packet, until it is received by the other swarm members. In this work we refer to an iteration also as a time slot $t_s$.
$A_p$ is maintained by each agent independently based on information received by others.
Considering a periodic time slot assignment like in TDMA, the update time for DTSA, $\bar t_{u,a}$, can be defined as:
\begin{equation}
    \bar t_{u,a} = N_{p} \, {t_{s}}, \quad \quad N_p \le N
    \label{eq:update_time_NA}
\end{equation}
where, $N_p = |A_p|$. This formulation is more adaptive and appropriate for situations where we know that a part of the swarm may not be generating new information.
From the previous example in Section \ref{section:system model}, now our $N_{p} = 2$, which reduces the update time for DTSA by half and the flying agents can get updated information from each other faster.

In DTSA each agent $a$ keeps estimating how far it is from its target position, $\mathbf{p}_{\text{target}}(a)$, using: 
\begin{align}
    \mathbf{p}_e^a = \| \mathbf{p}_{\text{target}}(a) - \mathbf{p}_i(a) \|_{2}.
\end{align}
where, $\mathbf{p}_i(a)$ is the position of agent $a$ at iteration $i$.
Agent $a$ keeps using its allocated time slots as long as $\mathbf{p}_e^a > r_{\text{min}}$, where $r_{\text{min}}$ is a predefined distance.
This is done to avoid sending redundant information once the agent is at its destination.
So, in the allocated time slot, along with its velocity, $\mathbf{v}(a)$, and position, $\mathbf{p}(a)$, the agent also broadcasts a transmission parameter $c^a_{r}$ which is defined as:
\begin{equation}
    c^a_{r} = 
    \begin{cases}
	C,  & \mathbf{p}_e^a > r_{\text{min}} \\
	0, & \mathbf{p}_e^a \leq r_{\text{min}} \hspace{5pt}\\
	\end{cases} 
	\label{eq:remaining_t_slots}
\end{equation}
where, $C$ is a predefined constant used by all agents.
As soon as $\mathbf{p}_e^a \leq r_{\text{min}}$, agent $a$ transmits $c^a_{r} = 0$. On receiving this information, the other agents can remove agent $a$ from their set $A_p$.
This is useful since some agents can reach their target before the others, or they may need to leave the swarm, e.g., due to low battery or other malfunction.

Another objective of the DTSA is to dynamically assign time slots, to be adaptive to the flight behaviour of other agents.
Since the next sender, $a_s$, can be any agent in the set $A_p$, the algorithm is no longer periodic.
One idea could be to select the $n$ senders for the next $n$ iterations but not only does that require high computational power, it is also not very adaptive to sudden changes in the flight behaviours. 
Therefore, at each iteration $i$ we select a new sender for the iteration $i+1$.
In order to select the next sender $a_s$, each agent $a \in A$, first calculates a priority value, $g^{a}_{k}$, for each agent $k \in A_p$, as explained next. 

\subsection{Priority Value}
Every agent $a \in A$ calculates the priority of every agent $k \in A_p$, w.r.t. all agents in the swarm $j \in A$. 
And the agent with the highest priority is allocated the next time slot.
In order to calculate this priority value we use various parameters which describe the flight behaviour of the agents.
One parameter that we use 
is the relative velocity between the $j$th and $k$th agent, $j \neq k$, which is defined as
\begin{equation}
\mathbf{v}_{j,k} = \mathbf{v}_{j} - \mathbf{v}_{k} = \frac{\text{d}\mathbf{p}_{j}}{\text{d}t} - \frac{\text{d}\mathbf{p}_{k}}{\text{d}t},
\end{equation}
where, $\mathbf{p}_{j}$ and
$\mathbf{p}_{k}$ are the position vector of the $j$th and $k$th agent, respectively. The relative velocity of two agents tells us how fast and in which direction these agents are flying, relative to each other. For a small enough time slot length $t_s$, we can obtain the distance that the agents cover relative to each other, which is the euclidean norm of the relative displacement $\mathbf{d}_{j,k}$  calculated as
\begin{equation}
     \|\mathbf{d}_{j,k}\|_{2} = \|\mathbf{v}_{j,k}\|_{2}\cdot{t_{s}} 
\end{equation}
The relative displacement is a measure of how far two agents are moving relative to each other during the time slot $t_s$. The idea is that agents that cover large distances in a short period of time can cause collisions, and also their last sent position is not accurate anymore, which means that they must send their new position. 
However, the relative displacement is not enough to determine how prone to collision are the movements of the two agents. For example, consider two different agents $k$ and $j$ which are flying in a straight line towards each other
with a constant relative velocity of $0.3\frac{\text{m}}{t_s}$. If they are at a distance of $3$m from each other, then it means that the agents still have $10$ iterations before they collide. But if the distance is only
$0.6$m then they will collide after 2 iterations. 

Hence, we have to take the distance between the agents into account. In order to describe this behavior, we use the ratio between the euclidean norm of the relative displacement and the distance between the agents,  to obtain  $\frac{\|\mathbf{d}_{j,k}\|_{2}}{\|\mathbf{p}_{k,j}\|_{2}},$

where $\|\mathbf{p}_{k,j}\|_{2}$ is the euclidean distance between two agents, $k$ and  $ j$, and is calculated with 
\begin{equation}
    \|\mathbf{p}_{k,j}\|_2 = \|\mathbf{p}_{k} - \mathbf{p}_{j}\|_2,
\end{equation}
where, $\mathbf{p}_{k,j}$ is a vector from the $j$th agent to the $k$th agent that describes the position of agent $k$ relative to the position of the  agent $j$. The ratio  $    \frac{\|\mathbf{d}_{j,k}\|_{2}}{\|\mathbf{p}_{k,j}\|_{2}}$ tells us how much distance is covered by the agents in one iteration, relative to the distance between them. 
For example,  $ \frac{\|\mathbf{d}_{j,k}\|_{2}}{\|\mathbf{p}_{k,j}\|_{2}} > 1$ tells us that the distance of the relative displacement of the agents is greater than their distance $\|\mathbf{p}_{k,j}\|_{2}$, but we do not know in which direction the agents are flying.  
If, with $ \frac{\|\mathbf{d}_{j,k}\|_{2}}{\|\mathbf{p}_{k,j}\|_{2}} > 1,$ they are flying in a straight line towards each other, collision will occur in the next iteration, but if they fly in another direction there will be no collision. 
Therefore, we have to consider the angle, $\alpha_{j,k}$, between the relative velocity and relative position of the $k$th and $j$th agents, in order to determine if the movement is prone to collision.
We determine this angle as

\begin{align}
   \alpha_{j,k} = \arccos{\frac{\mathbf{v}_{j,k} \cdot \mathbf{p}_{k,j}}{\|\mathbf{v}_{j,k}\|_{2}\cdot \|\mathbf{p}_{k,j}\|_{2}}}, \quad \alpha_{j,k} \in [0, \pi]
   \label{eq:permission_alpha}
\end{align}
An angle $\alpha_{j,k} = 0$, means that the agents are flying towards each other. In this case we want the priority value of the $k$th agent to take a higher value, compared to a situation with the same relative velocity and distance but with an angle  $\alpha_{j,k} > 0$.  
Therefore, to give higher value to smaller angles we introduce the normalising term 
\begin{equation}
    \frac{\pi - \alpha_{j,k}}{\pi} \in [0,1],
\end{equation}
In order to obtain a combined term that describes the flight behavior of the $j$th agent relative to the flight behavior of the $k$th agent we use

\begin{equation}
   \eta_{j,k} =  \frac{\|\mathbf{d}_{j,k}\|_{2}}{\|\mathbf{p}_{k,j}\|_{2}}\cdot{ \frac{\pi - \alpha_{j,k}}{\pi}}.
\end{equation}
where, $\eta_{j,k} $ only describes relative behaviour only between two agents, $j,k$. 
And to describe the movement of each agent of the swarm relative to the $k$th agent, we determine $\eta_{j,k} $ for all $j \in A$, and sum up these terms, to obtain our priority value at iteration $i$, $g^{a}_{k}$  for the $k$th agent, given as


\begin{align}
g^{a}_{k}\nonumber
&= \sum_{j \in A, j \neq k}\frac{\|\mathbf{v}_{j,k}\|_{2} \, {t_{s}}}{\|\mathbf{p}_{k,j}\|_{2}} {\frac{\pi - \alpha_{j,k}}{\pi}}\\
\nonumber&= \sum_{j \in A, j \neq k}\frac{\|\mathbf{d}_{j,k}\|_{2}}{\|\mathbf{p}_{k,j}\|_{2}} {\frac{\pi - \alpha_{j,k}}{\pi}}\\
&= \sum_{j \in A, j \neq k}\eta_{j,k} 
\label{eq:permission_d_p_ang}
\end{align}
The index $a$ indicates that each agent $a \in A$ performs this calculation on its own, in a decentralized manner.
In order to give a complete description of the priority value $g^{a}_{k}$, we have to define how we treat individual components of this function. 
Note that if $\|\mathbf{p}_{k,j}\|_{2} = 0$, this means that a collision has occurred and the task has failed and is terminated, so we do not consider this case. And if $\|\mathbf{v}_{j,k}\|_{2} = 0$ this
means that the agents retain the same distance between each other while flying in the same direction, which is highly unlikely for moving quadcopters. 
However, if this does occur we set $\eta_{j,k}  = 0$ because their relative velocity is $0$ at this point.
For the special case where $\|\mathbf{v}_{k}\|_{2} = 0$, the priority value of the $k$th agent is set to be $0$, because the $k$th agent does not change its position. This is suitable for situations, in which the $k$th agent has to stop for a while during the operation or if it has reached its target position.
Once every agent $a$ has calculated the priority values for all agents, it then uses the selection function, explained in the next section, to select  the next sender. 

\subsection{Selection Function}
After every agent has calculated the priority values vector $\mathbf{g}^a_{k}$ for all other agents, it then makes use of a comparison value, $w^{a}_{k,j}$, to see who has the highest priority.
$w^a_{k,j}$ is defined as:

\begin{align}
w^{a}_{k,j} = \frac{g^{a}_{k} - g^{a}_{j}}{g^{a}_{k}} > \epsilon
\label{eq:pfs}
\end{align}
where, $k, j \in A_p$, $k \neq j$, and $\epsilon$ is the minimum threshold to be considered higher priority than the other agents. 
Using $w^a_{k,j}$ and $\epsilon$ the algorithm attempts to determine if the priority values of two agents are significantly different from each other.
We do element wise comparison of the priority values and always keep the greater value $g^{a}_{k}$, so we assume that $g^{a}_{k} > g^{a}_{j}$.
Once all the comparisons are done, you end up with the agent $k$ which has the highest $g^{a}_{k}$, which is then selected.

Every agent $a$ also maintains a counter, $h^a_{i,k}$, for itself and every other agent, which keeps track of the number of time slots since each agent last transmitted.
The counter is defined as follows:
\begin{equation}
    h^{a}_{i,k} = 
    \begin{cases}
	0, & k = a_s \\
	h^{a}_{i-1, k}+1, & k \neq a_s, \hspace{5pt}\\
	\end{cases} 
	\label{eq:secondary}
\end{equation}

This is useful to break ties in the priority values and also to avoid any agents' information from being too outdated.
\subsection{Data Estimation}
\label{subsection:estimation}
\begin{figure}  
    \centering
    \includegraphics[width=.48\textwidth]{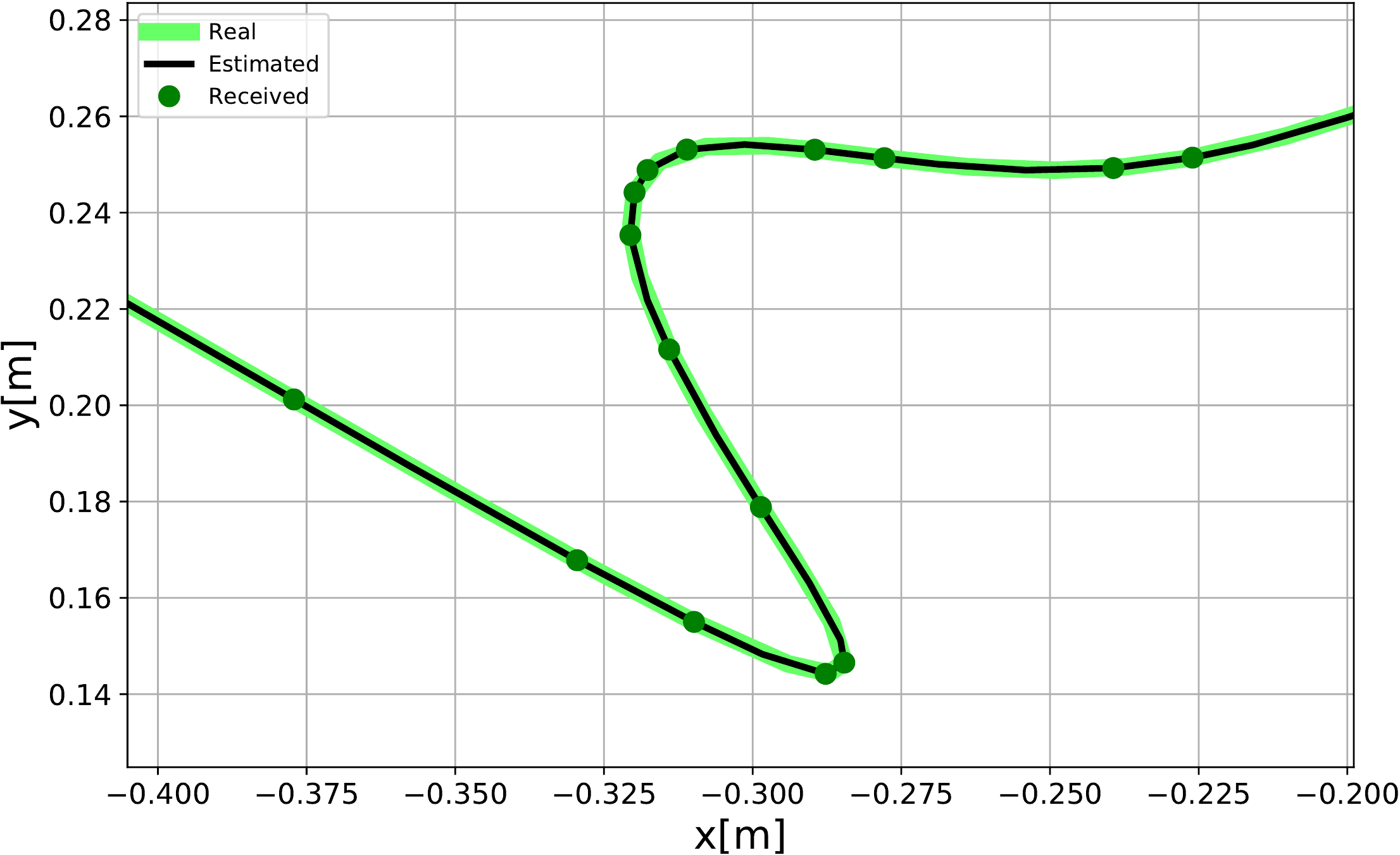}
    \caption{Trajectory comparison of a quadcopter. The real trajectory (green) is the one this quadcopter actually took, while the estimated trajectory (black) is the one estimated by the other quadcopter. Green  dots  are  the received position values by the estimating quadcopter. Negative axis because we consider the center of the room to be located at $(x=0,y=0,z=0)$.}
    \label{fig:estimated}
\end{figure} 
In order to evaluate the priority values in a decentralised manner, each agent needs to maintain an estimate of the position of all other agents, except the last sender $a_s$ since we already have the updated information for that one. 
This estimation is done using:
\begin{align}
    \mathbf{p}^{a}_{i,k}  &=
    \mathbf{p}^{a}_{m,k} + (i-m)\mathbf{v}^{a}_{m,k}\,{t_{s}}
    \label{eq:estimate}
\end{align}
where, $m < i$, is the last iteration in which the $k$th agent shared its velocity, $\mathbf{v}_k$, and position, $\mathbf{p}_k$.
Note that each agent also maintains an estimate of its own position, even though it has access to its actual position at all times. This is done to keep the estimates consistent over all agents, for a decentralised but unanimous comparison.
This estimated updated position is then also used in the collision avoidance algorithm.
In Fig. \ref{fig:estimated} we show a small part of the trajectory of a quadcopter.
The real trajectory (green) is the one this quadcopter actually took, while the estimated trajectory (black) is the one estimated by the other quadcopter, using Eq. \ref{eq:estimate}.
The green dots are the received position values by the estimating quadcopter. 
It can be seen that this linear estimator is able to predict the trajectory very well, helping greatly in avoiding collision.

\section{RESULTS}
\label{section:results}


In this section we will first explain the general setup for our quadcopters and then give our results first for the simulations and then the experiments on real quadcopters.
We assume that the agents can communicate without using a central authority (decentralised) and their message is broadcasted to everyone in their communication range. 
The time slot length, $t_{s}$, is fixed and is defined before starting the task. The id of each quadcopter is assigned initially and is fixed.
We do not use acknowledgment packets, and unlike traditional TDMA algorithms, DTSA selects only the sender for the next time slot, $t_{s}$. 
For TDMA, the agents, in their respective time slots, broadcast their id, position and velocity. 
While, in DTSA they additionally communicate their  transmission parameter, $c^a_r$ (Eq. \ref{eq:remaining_t_slots}).

The task for the quadcopters is to switch positions with the quadcopters opposite them in a circle or an ellipsoid. 
The target co-ordinates are initially communicated to all the quadcopters.
Each agent uses a decentralised collision avoidance algorithm, where in DTSA they make use of their position estimates, explained in Section \ref{subsection:estimation}, to predict and avoid other agents.
We will not go into details of the collision avoidance algorithm since it was not the focus of this work.
DTSA can be used in combination with any collision avoidance algorithm and any formation change application.
\textit{Note that once a collision occurs we assume that the task was unsuccessful and do not consider its result while plotting.}

The parameters used in this work to compare the performance of TDMA and DTSA are:
\begin{itemize}
    \item Minimum distance: the minimum euclidean distance between the quadcopters at any point during the task. The closer they get, higher is the chance of collisions.
    \item Trajectory efficiency: the ratio between the minimum distance quadcopter should have travelled (straight line) and total distance travelled by each quadcopter. 
    \item Completion time: the time taken by all the quadcopters to exchange their positions (considered only for real quadcopters).
\end{itemize}

In order to highlight the need for dynamic time slot allocation we use three different setups. One setup is in which all $N$ quadcopters are moving, second is when half of the quadcopters (or 10 out of $N=18$ quadcopters) are moving and lastly when only $2$ out of $N$ quadcopters are moving. The quadcopters which are not moving are still part of the task but just hovering at their initial positions, so they do not have any new information to transmit. For all the results we use $\epsilon=0.5$ in Eq. \ref{eq:pfs}, and $r_{\text{min}}=0.3$m in Eq. \ref{eq:remaining_t_slots}, unless stated otherwise. We performed $10$ Monte Carlo simulations for each setup.

\begin{figure}
    \centering
    \includegraphics[width=.48\textwidth]{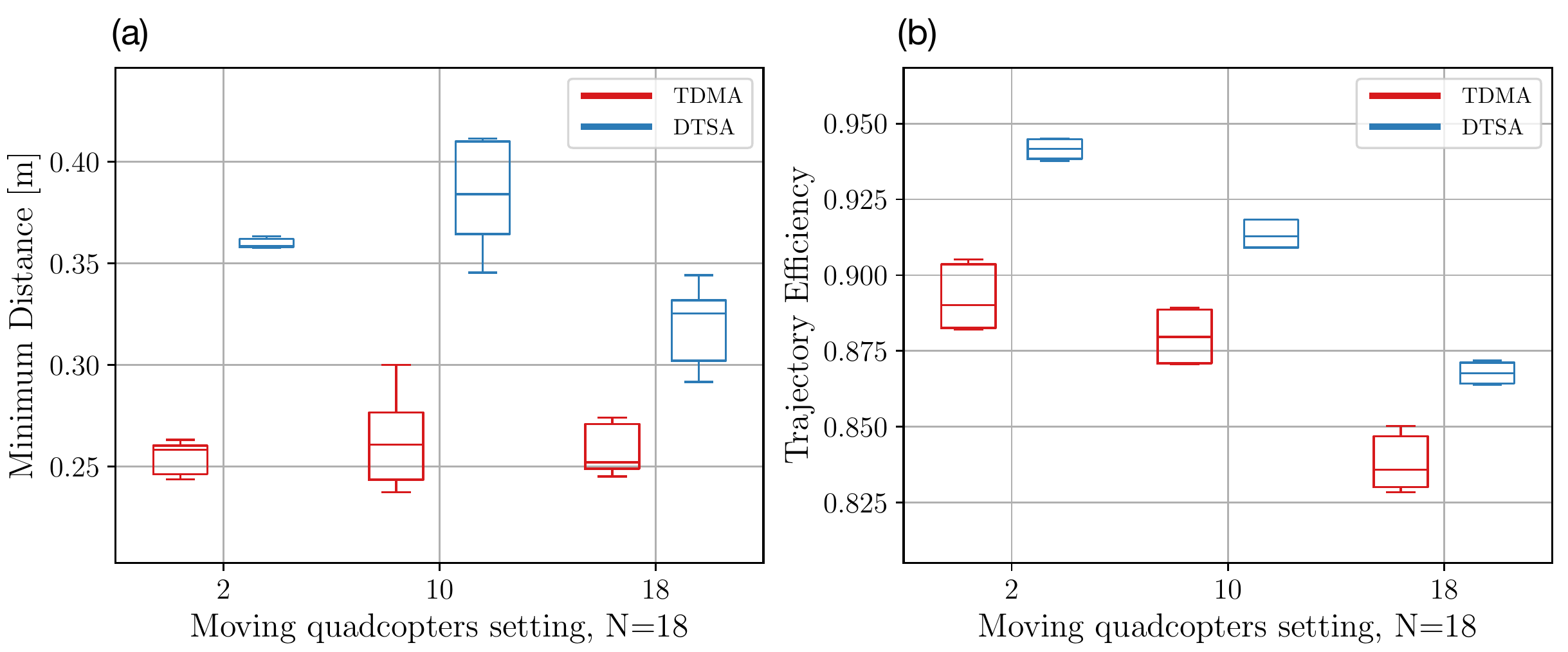}
    \caption{Simulation results for $18$ quadcopters are shown. In each sub figure, the x-axis represents the number of moving quadcopters out of $18$. It can be seen that DTSA outperforms TDMA in terms of minimum distance and trajectory efficiency for all three of moving setups.}
   \label{fig:18_sim}
\end{figure}


\paragraph*{Where TDMA fails?} Before going into the results we want to highlight conditions in which TDMA failed to avoid collisions, resulting in our chosen parameters for the following results. 
Firstly, due to the allocation of time slots to all $N$ quadcopters, the number of drones successfully working is limited in TDMA.
This is because the more quadcopters you have the higher will be the update time for each, resulting in longer outdated information and eventually collisions.
From our simulations we found that when all $N$ agents are moving, for $N > 18$ TDMA always has collisions, since the update time, Eq. \ref{eq:update_time_tdma} is too high.
Whereas, DTSA works without collisions for upto $N=70$ drones.
Due to this limitation, we will restrict ourselves to number of drones where TDMA also works successfully, so that we can have a performance comparison.
Secondly, if we increase the time slot length, $t_s$ from $10$ms to $20$ms, then this reduces the TDMA to $N=10$ quadcopters, while with DTSA we can still work with $N=60$ quadcopters. 
Due to lack of space we do not give performance results for this scenario.

In the following, first we show the simulation results and then experiment results with real Crazyflie quadcopters.

\subsection{Simulations}
\label{subsection:simulations}
In our simulations we consider the quadcopters as point masses, so we assume that if the minimum distance between them gets less than $0.2$m there is a collision. 
For the simulations we also do not consider the battery life, packet size or other physical aspects of the environments which can affect the real quadcopter experiments, like wind or disturbance due to other nearby quadcopters. The simulation setup is written in Python programming language, using the Panda3D \cite{pd3d} game engine. 


\begin{figure}
    \centering
    \includegraphics[width=.48\textwidth]{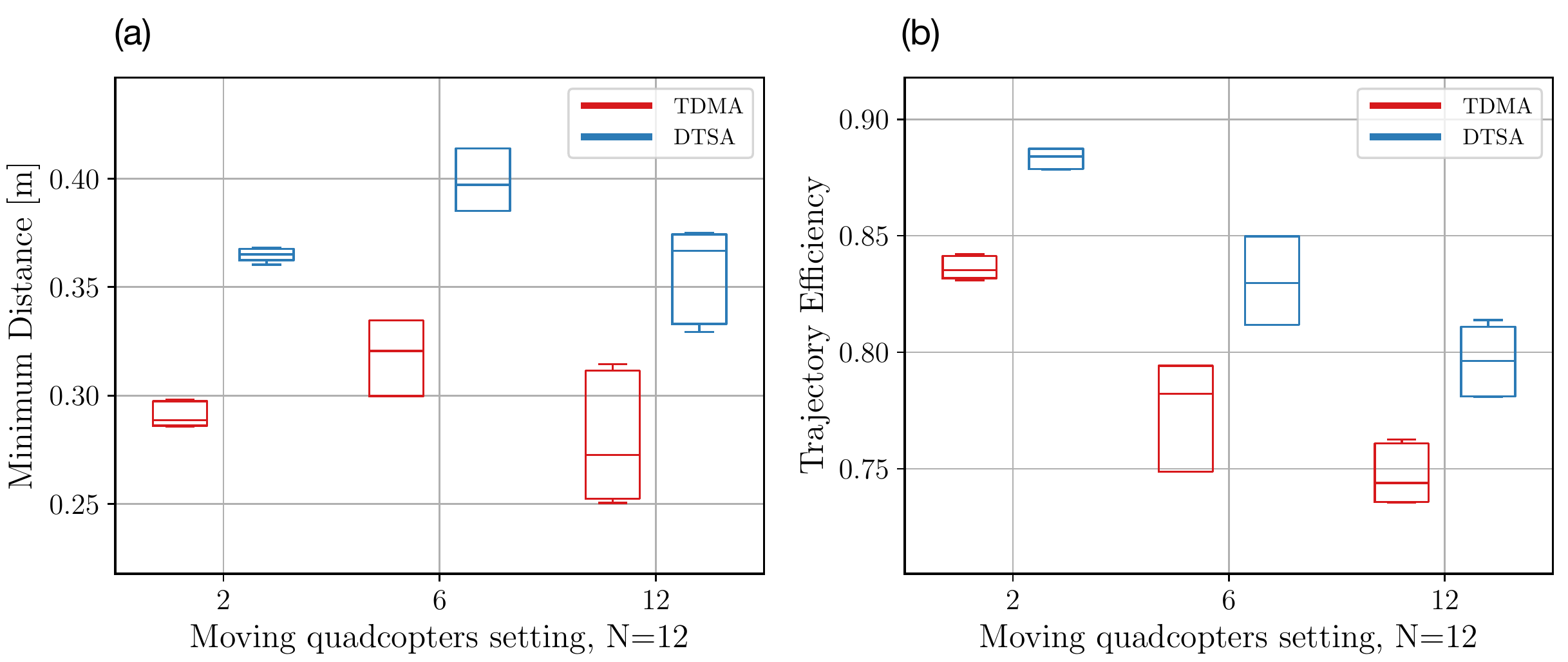}
    \caption{Simulation results for $12$ quadcopters are shown. In each sub figure, the x-axis represents the number of moving quadcopters out of $12$. It can be seen that DTSA outperforms TDMA in terms of minimum distance and trajectory efficiency for all three of moving setups.}
    \label{fig:12_sim}
\end{figure} 


In Fig. \ref{fig:18_sim} we have $N=18$ quadcopters for all three moving setups. 
It can be seen in Fig. \ref{fig:18_sim}(a) that due to longer outdated information in TDMA the quadcopters get too close to each other, before realising the presence of the other and then have to move away abruptly, leading to worse trajectories, as seen in Fig. \ref{fig:18_sim}(b).
In Fig. \ref{fig:12_sim} we have $N=12$ quadcopters for all three moving setups. 
The performance trend is the same as in Fig. \ref{fig:18_sim}.
We do this simulation to compare later to our experiments, which we also do with $N=12$ Crazyflie quadcopters.

\subsection{Experiments}
\label{subsection:experiments}

 \begin{figure*}
    \centering
    \includegraphics[width=.98\textwidth]{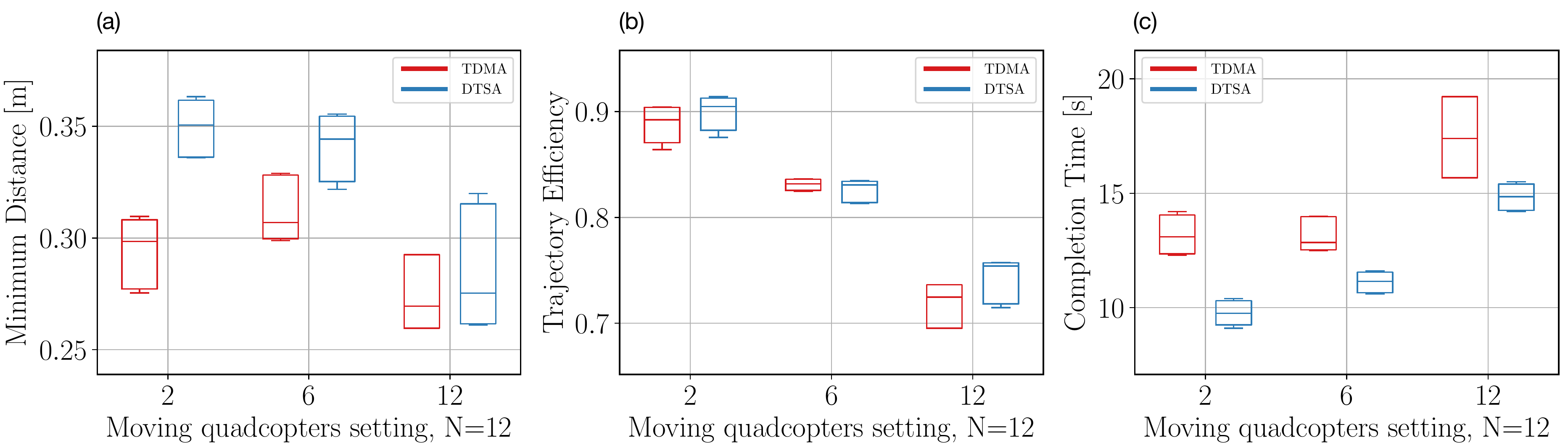}
    \caption{Experiment results, performed in our Self-Organizing Systems Lab, for $12$ Crazyflie quadcopters are shown. In each sub figure, the x-axis represents the number of moving quadcopters out of $12$. It can be seen that DTSA out performs TDMA in terms of minimum distance, trajectory efficiency and completion time.}
    \label{fig:12_exp}
\end{figure*} 
We did the experiments in our lab in a room of the size $x=3.0$m $\times$ $y=4.6$m $\times$ $z=1.9$m using Crazyflie quadcopters. 
Crazyflie are nano quadcopters provided by Bitcraze, equipped with four $7$mm $\times 16$mm coreless DC motors and $45$mm plastic propellers.
Each Crazyflie is equipped with a lighthouse expansion deck which allows it to estimate its own global position and is being controlled over the $2.4$GHz Crazyradio PA (with Nordic Semiconductor nRF24LU1+) in up to $1$km range line-of-sight  (with  transmission  up  to  $2$Mb/sec  in  $32$-byte  packets) \cite{crazyflie}. We synchronize the clocks of the quadcopters at the start of the experiments.
Due to the limitation of the size of our room we were only able to perform experiments with $N=12$ Crazyflie quadcopters.
Higher number of quadcopters than this could not be achieved because with too many quadcopters downwash was too high in this small area, leading to unstable flights.

\begin{figure}
    \centering
     \includegraphics[width=.48\textwidth]{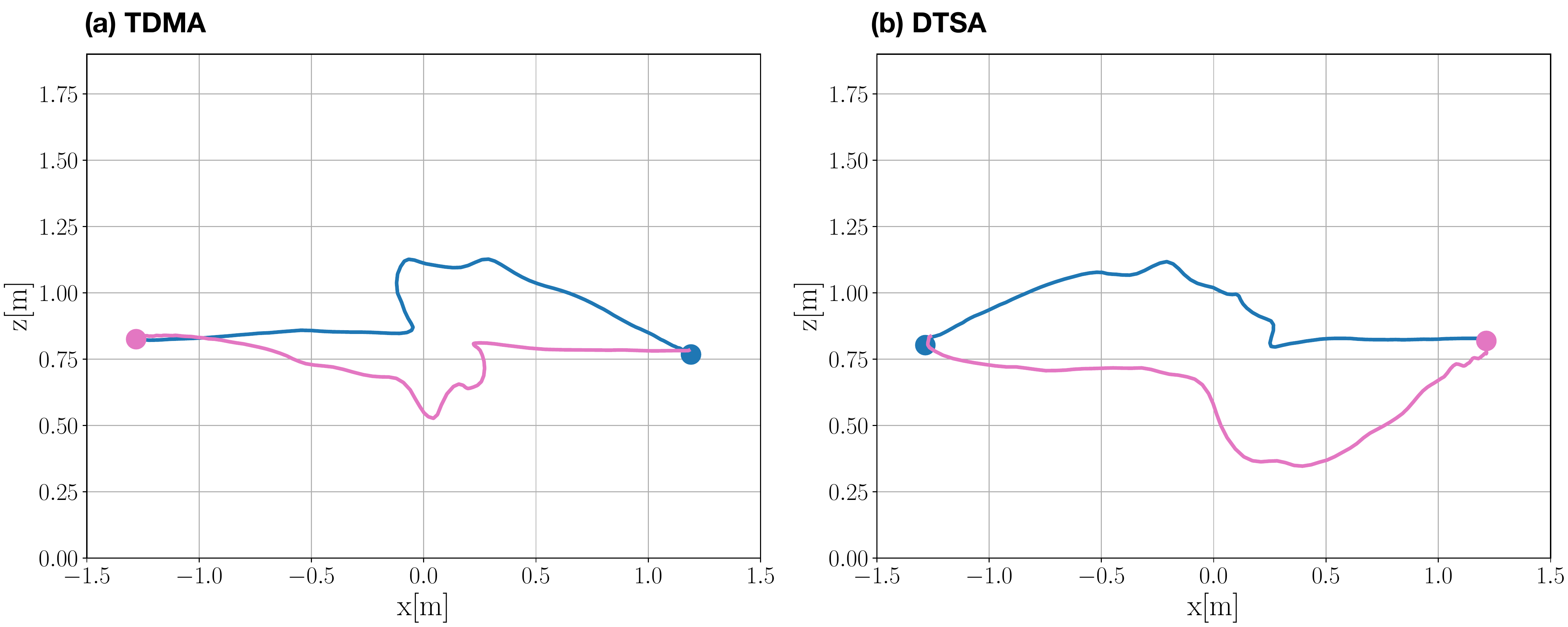}
    \caption{Trajectory comparison of $2$ moving quadcopters out of $12$. DTSA has a much smoother trajectory than TDMA resulting in smaller completion time and better trajectory efficiency. Negative axis because we consider the center of the room to be located at $(x=0,y=0,z=0)$.}
    \label{fig:12_exp_traj}
\end{figure}

In Fig. \ref{fig:12_exp} we have a similar setup as in Fig. \ref{fig:12_sim} but now with real quadcopters. It can be seen here as well that DTSA has better performance in terms of minimum distance Fig. \ref{fig:12_exp}(a) and trajectory efficiency Fig. \ref{fig:12_exp}(b). This is because the agents using DTSA are more aware of the positions of the other agents and are able to plan their trajectories smoothly, which then also leads to a smaller completion time, see Fig. \ref{fig:12_exp}(c). While, in TDMA it was observed that all agents come first towards the center, trying to take the shortest path, and then try to avoid collisions, resulting in a jerky motion, because they do not have the correct position of each other. This might lead to higher trajectory efficiency for smaller number of moving quadcopters but as the number of moving quadcopters increases this approach is no more efficient, as can be seen in Fig. \ref{fig:12_exp}(b). 

Fig. \ref{fig:12_exp_traj} shows the trajectory of $2$ Crazyflie quadcopters. In Fig. \ref{fig:12_exp_traj}(a) the agents use TDMA and in Fig. \ref{fig:12_exp_traj}(b) they use DTSA. It can be seen that the path taken by DTSA is much smoother resulting in higher trajectory efficiency and lower completion time, specially as the number of agents will increase.

\section{CONCLUSION}
\label{section:conclusion}
In this work we present a new decentralized channel access protocol for the quadcopters, dynamic time slot allocation, DTSA, which enables the agents to be adaptive and hence perform better than the state-of-the-art division multiple access scheme, TDMA. 
Our proposed algorithm can be used in scenarios where agents do not have new information to communicate, or if they want to leave the task, or are dead because of battery or other malfunction.
We show the enhanced performance due to dynamic allocation not only in simulations but also with Crazyflie quadcopters provided by Bitcraze.
One future direction will be to enhance the adaptivity such that it allows agents to also enter in between the task. 



\bibliographystyle{IEEEtran}

\bibliography{biblio}

\end{document}